\documentclass{article}
\usepackage[T1]{fontenc}
\usepackage{iclr2027_conference,times}
\iclrfinalcopy

\usepackage{amsmath,amssymb,amsthm}
\usepackage{graphicx}
\usepackage{booktabs}
\usepackage{multirow}
\usepackage{array}
\usepackage{xcolor}
\usepackage{colortbl}
\usepackage[most]{tcolorbox}
\usepackage{microtype}
\usepackage{caption}
\usepackage{hyperref}
\usepackage{url}
\usepackage[capitalize,noabbrev]{cleveref}

\definecolor{pnmark}{RGB}{142,90,168}
\definecolor{vblue}{RGB}{46,147,196}
\definecolor{citeblue}{rgb}{0.21,0.49,0.74}
\definecolor{myskyblue}{RGB}{135, 206, 235} 
\definecolor{myorange}{HTML}{FF8C00}
\definecolor{oursgray}{HTML}{EFEFEF}

\hypersetup{
  colorlinks=true,linkcolor=myorange,citecolor=citeblue,urlcolor=vblue,
  pdftitle={Perturbot: Breaking Shortcut Priors in Vision-Language-Action Models with Perturbative Training},
  pdfauthor={Mingyu Liu, Chonghao Sima, Tianjian Feng, Hanqing Wang, Cong Chen, Hao Chen, Chunhua Shen}
}

\makeatletter
\let\pb@NAT@parse\NAT@parse
\renewcommand\NAT@parse[1]{%
  \pb@NAT@parse{#1}%
  \ifx\NAT@date\@empty\else
    \xdef\NAT@date{{\noexpand\color{myorange}\unexpanded\expandafter{\NAT@date}}}%
  \fi}
\makeatother
\newif\ifdraftnums
\draftnumsfalse

\ifdraftnums
  
\else
  
\fi
\newcommand{\mn}[1]{#1}

\newcommand{\cim}[1]{\,{\scriptsize$\pm$#1}}

\newcommand{\vcap}{\emph{salience capture}}

\newcommand{\llock}{\emph{noun lock-in}}

\newcommand{\minert}{\emph{motor inertia}}

\newcommand{\method}{\mbox{Perturbot}}
\newcommand{\metric}{\mbox{GroundFscore}}
\newcommand{\gf}{\mathrm{GF}}

\newcommand{\chan}{c}

\newcommand{\chanV}{c_{V}}
\newcommand{\chanL}{c_{L}}
\newcommand{\chanA}{c_{A}}

\newcommand{\obs}{o}
\newcommand{\vision}{v}
\newcommand{\third}{\vision^{\mathrm{third}}}
\newcommand{\wrist}{\vision^{\mathrm{wrist}}}
\newcommand{\instr}{\ell}
\newcommand{\prop}{q}
\newcommand{\act}{a}
\newcommand{\pol}{\pi_{\theta}}

\newcommand{\TV}{T_{V}}

\newcommand{\TC}{T_{C}}

\newcommand{\TR}{T_{R}}

\newcommand{\Enull}[1]{\mathcal{E}^{0}_{#1}}
\newcommand{\Ecaus}[1]{\mathcal{E}^{1}_{#1}}

\DeclareMathOperator{\MI}{I}
\DeclareMathOperator{\Ent}{H}

\newcommand{\data}{\mathcal{D}}

\newcommand{\mshortcuts}{\emph{modality shortcuts}}

\newcommand{\up}{$\uparrow$}
\newcommand{\down}{$\downarrow$}

\newcommand{\oursrow}{\rowcolor{oursgray}}

\title{{\includegraphics[height=1.25em]{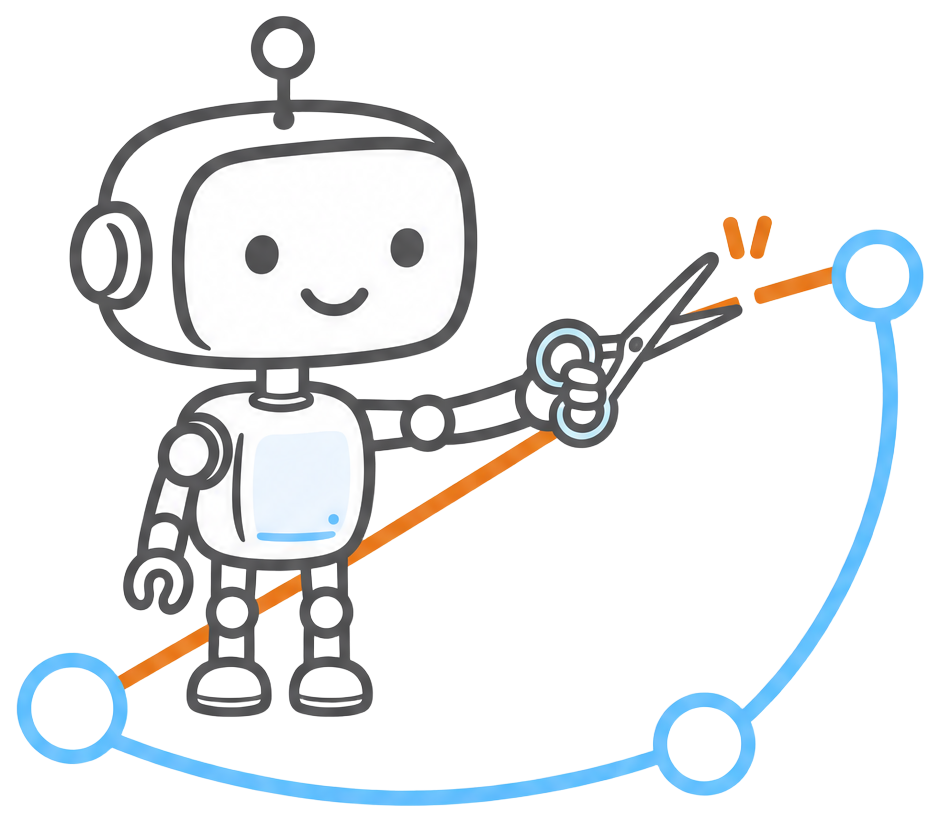}} \textcolor{myorange}{Pertur}\textcolor{myskyblue}{bot}: Breaking Shortcut Priors in Vision-Language-Action Models
with Perturbative Training}

\author{\parbox[t]{\dimexpr\textwidth-2\tabcolsep\relax}{\centering\bfseries
Mingyu Liu\textsuperscript{1,2,*}\quad Chonghao Sima\textsuperscript{3,*}\quad
Tianjian Feng\textsuperscript{1}\quad Hanqing Wang\textsuperscript{4}\\
Cong Chen\textsuperscript{1}\quad Hao Chen\textsuperscript{1,\ensuremath{\dagger}}\quad
Chunhua Shen\textsuperscript{1,\ensuremath{\dagger}}\\[0.4em]
{\normalfont\small
\textsuperscript{1}Zhejiang University\quad
\textsuperscript{2}Shanghai Innovation Institute\quad
\textsuperscript{3}University of Hong Kong\quad
\textsuperscript{4}HKUST(GZ)}\\[0.3em]
  \small{\href{https://github.com/aim-uofa/PerturBot}{{\color{orange} \textbf{github.com/aim-uofa/PerturBot}}}}
  \vspace{-1.5em}
  }}
\date{}

\begin{document}
\maketitle
\lhead{Preprint}
\begingroup
\renewcommand{\thefootnote}{\fnsymbol{footnote}}
\footnotetext[1]{Equal contribution.
\textsuperscript{\ensuremath{\dagger}}Corresponding authors.}
\endgroup

\begin{abstract}
A vision--language--action (VLA) policy can complete complex tasks while ignoring the evidence that should determine its actions. An object held near the wrist camera can displace the instructed target. Language and action show the same pattern: a familiar noun can trigger the operation it was paired with in training even after the verb changes, and a gripper that closed on nothing may lift anyway. We call these dependencies \mshortcuts{}: regularities in successful demonstrations make visual, lexical, or motor cues sufficient to predict expert actions without the task evidence needed for the underlying decision. More demonstrations of the same kind can raise task success while leaving these shortcuts intact. We propose \method{}, which makes task-relevant evidence easier to use and shortcuts insufficient on their own: it applies task-preserving wrist-view perturbations, enriches instructions with decision-relevant captions, and adds random and failed trajectory segments relabeled with the behavior they contain. It complements scaling by changing what is scaled, and leaves inference unchanged. Moreover, we propose \metric{}, an offline score that diagnoses how severely a policy relies on modality shortcuts. Task success rate shows whether a policy improves, while \metric{} reveals whether the policy scales healthily, relying on task evidence rather than shortcuts. Together, \method{} and \metric{} provide a training-and-evaluation framework for disentangling VLA decisions from shortcut priors while preserving responsiveness to task-relevant evidence.

\end{abstract}

\section{Introduction}
\label{sec:intro}

\begin{figure}[t]
\centering
\includegraphics[width=0.95\linewidth]{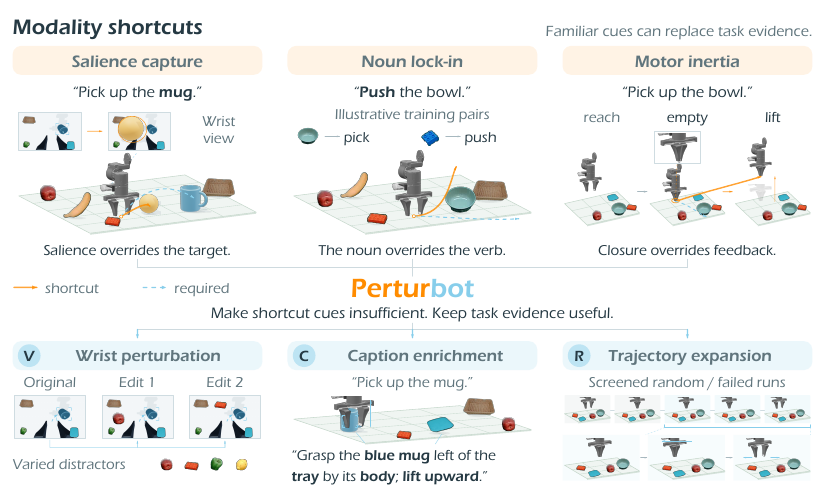}
\caption{Modality shortcuts and \method{}. Top: familiar visual, lexical, and
motor cues can replace task evidence (solid: hypothetical shortcut choices;
dashed: required alternatives). Bottom: task-preserving wrist-view
perturbations (V), behavior-supported caption enrichment (C), and screened,
relabeled trajectory segments (R; \cref{sec:training}). Examples are conceptual;
relabeled local behavior does not by itself establish recovery.}
\label{fig:overview}
\end{figure}

\begin{flushright}
\emph{``Success is a lousy teacher.''}\\
{\small --- Bill Gates, \emph{The Road Ahead} (1995)}
\end{flushright}
Vision--language--action (VLA) policies can now perform complex manipulation
tasks \citep{zitkovich2023rt2,kim2024openvla,ghosh2024octo,black2025pi0}, yet a
simple change in view can still divert them (\cref{fig:overview}). An object
held near the wrist camera can displace the instructed target; changing ``pick
up the bowl'' to ``push the bowl'' may still elicit a grasp; and a gripper that
closes on nothing may lift anyway. Task success cannot
tell whether the policy uses the scene, instruction, and observed outcome to
decide what to do, or merely replays an action when a familiar cue appears.

We call such dependencies \mshortcuts{}, a multimodal instance of shortcut
learning \citep{geirhos2020shortcut} closely related to causal confusion in
imitation learning \citep{dehaan2019causal,wen2020fighting}. A modality shortcut
is not an unimportant input but an easily decoded cue that stands in for the
evidence a decision requires. Such cues arise from the regularity of successful
demonstrations: the target dominates the wrist view as the expert reaches for
it, each object name may be paired with a single operation, and gripper closure
is reliably followed by a lift. Behavior cloning rewards agreement with the
demonstrated action, not the evidence used to predict it, so nothing in the
objective stops the policy from relying on the cue.

This matters for scaling. Larger and more diverse demonstration corpora are
central to generalist robot learning
\citep{oneill2024open,khazatsky2024droid,lin2025scaling}, but if new
trajectories preserve the same correlations, the shortcut rule and the intended
rule keep prescribing the same actions, and more data cannot distinguish them.
Higher success on familiar tasks then need not mean more reliable target
selection, instruction following, or outcome assessment: the policy grows more
proficient where shortcut and evidence agree yet stays brittle where they
disagree, which makes modality shortcuts a bottleneck for scaling.

Addressing this requires changing what the training data demand, not weakening
an input modality: the evidence a correct decision needs should be easier to
use, and shortcuts should be insufficient on their own (\cref{sec:analysis}).
\method{} does so with three training-time data interventions and no extra model or
forward pass at deployment: task-preserving wrist-view perturbations add
plausible distractors without changing the target or invalidating the recorded
action; decision-relevant captions make object attributes, spatial relations,
and requested operations explicit, each supported by the recorded behavior; and
caption-grounded trajectory expansion adds random-motion and failed-execution
segments relabeled with the behavior they contain, which add behavioral support
rather than recovery skills, since relabeling cannot create an unrecorded
correction. \method{} thus
changes what is scaled: as data grow, it expands not only repetitions of
familiar behavior but also the range of decisions that depend on the scene,
instruction, and observed outcome.

Whether scaling achieves this cannot be read from task success: when shortcut
and evidence agree, policies relying on either can succeed, and robustness to
distractors can come from simply ignoring vision. We therefore propose
\metric{}, an offline score that diagnoses how severely a policy relies on
modality shortcuts. It pairs each input with edits that should leave the action
unchanged, such as an off-path distractor, and edits that should change it in a
known way, such as naming a different object; a high score requires both, so
ignoring an input does not help. Because it needs no robot rollouts, \metric{}
can be computed for every checkpoint as data grow: task success shows whether a
policy improves, and \metric{} whether it improves for the right reason.

With a $\pi_{0.5}$ backbone, we evaluate all eight component combinations on 50
real-robot General PnP cases and cover broader skills in RoboTwin~2.0
\citep{chen2025robotwin2}; ablations test whether success and evidence use rise
together as data grow and whether \metric{} predicts online failures. Our
contributions are threefold:

\begin{enumerate}
\item \textbf{We identify \mshortcuts{} in VLA learning}, unifying \vcap{},
\llock{}, and \minert{}, and explain why scaling demonstrations that preserve
these correlations need not remove them.
\item \textbf{We propose \method{}}, three independently ablatable data
interventions that keep action supervision valid, change what is scaled, and
leave inference unchanged.
\item \textbf{We introduce \metric{}}, an offline score that, unlike task
success, indicates whether a policy scales healthily and cannot be raised by
ignoring useful evidence.
\end{enumerate}

\section{Related work}
\label{sec:related}

\textbf{Scaling VLA policies.}
Vision--language--action (VLA) policies map observations and instructions to
actions; they are trained on large robot datasets and often initialized from
pretrained vision--language models
\citep{brohan2023rt1,zitkovich2023rt2,kim2024openvla,ghosh2024octo,black2025pi0,black2025pi05,wang2025vq,liu2025gae,liu2026stamo}.
Their progress has followed the growth of demonstration data across tasks and
embodiments \citep{oneill2024open,khazatsky2024droid,walke2023bridgedata} and
studies of how data diversity and composition shape transfer
\citep{lin2025scaling,hejna2024remix,zha2025factored}, while evaluation has
expanded from nominal simulation benchmarks
\citep{liu2023libero,mees2022calvin,li2024evaluating} to perturbed scenes and
instructions \citep{fei2025liberoplus,zhou2025liberopro,zhao2026instructmove}.
Success alone cannot show whether a policy improves by using task evidence or by
using a shortcut more consistently. \metric{} measures shortcut reliance
offline; unlike action sensitivity to task-preserving perturbations, which has
been used to select demonstrations \citep{durso2026cfn}, it also requires the
action to change when task-relevant evidence changes.

\textbf{Shortcut learning.}
Neural networks often rely on features that are predictive in training but fail
under distribution shift, as studied under shortcut learning, simplicity bias,
and dataset bias \citep{geirhos2020shortcut,shah2020pitfalls,torralba2011unbiased}.
Visual question answering models can answer from language priors without
grounding the evidence in the image
\citep{goyal2017making,agrawal2018dont,cadene2019rubi}, and in imitation
learning, causal confusion and the copycat problem show that matching expert
actions need not identify the evidence that should control behavior
\citep{dehaan2019causal,wen2020fighting,park2021object}. Recent VLA studies find
that policies can underuse language or let vision override instructions
\citep{xu2025bayesvla,lian2026langforce,fang2026visionoverrides}, and diagnose or
regularize how modalities are used
\citep{shi2026vlatrace,xu2026unmasking,jia2026guidedvla,lin2026lit}. These
accounts usually examine one modality at a time; we tie each shortcut to the
decision it affects and trace all three to the regularity of successful
demonstrations.

\textbf{Data interventions for robust policies.}
Data-centric methods improve robustness by varying appearance through
augmentation and domain randomization
\citep{tobin2017domain,laskin2020reinforcement,bharadhwaj2023roboagent},
synthesizing or counterfactually editing demonstrations
\citep{mandlekar2023mimicgen,ameperosa2025rocoda}, enriching or relabeling
instructions \citep{kim2026demian,glossop2025cast}, and widening behavioral
coverage with corrective, perturbed, or unstructured data
\citep{ross2011reduction,laskey2017dart,lynch2019play,hu2025rac,liu2026perfect}. Perturbative
training has likewise reduced language-prior hallucinations in
vision--language models \citep{chen2025perturbollava}. Each of these
interventions is typically designed for one axis of variation, such as
appearance, geometry, language, or coverage; \method{} instead targets the cue
that substitutes for evidence in a given decision, while keeping every action
label valid and leaving inference unchanged.

\section{Method}
\label{sec:method}

We first model modality shortcuts (\cref{sec:analysis}), then describe the
three data components of \method{} (\cref{sec:training}) and how \metric{}
measures the resulting evidence use (\cref{sec:metric}).

\subsection{Modeling modality shortcuts}
\label{sec:analysis}
\label{sec:two_sources}
\label{sec:scaling}
\label{sec:phenomena}
\label{sec:failures}
We study three failures, one per decision a manipulation policy makes. In
\vcap{} (what to manipulate), an object near the wrist camera can divert a
reach from the instructed target; in \llock{} (which semantics apply), a policy
that has pushed other objects but only picked up the bowl may grasp it when
asked to push it; in \minert{} (how to move), a gripper that closed on nothing
may lift anyway. In each, a cue that is reliable in successful demonstrations
stands in for the required evidence: wrist-view salience for the instruction,
the noun for the verb, and the close--lift sequence for visual grasp feedback.

Consider a policy $\pol(\act\mid\vision,\instr,\prop)$ mapping camera
observations $\vision$ (wrist views included), an instruction $\instr$, and
motor context $\prop$ (proprioception, past actions) to an action or action
chunk $\act$. For a task decision $d$, a shortcut $s$, and task-relevant
evidence $e$, Bayes' rule under the data distribution $p$ separates a
shortcut-based prediction from an evidence update, whose expectation splits
again:
\begin{equation}
\label{eq:suffic}
p(d\mid s,e) = p(d\mid s)\,\frac{p(e\mid d,s)}{p(e\mid s)},
\qquad
\mathbb{E}\!\left[\log\frac{p(d\mid s,e)}{p(d\mid s)}\right]
= \MI(d;e\mid s) = \MI(d;e) - \mathcal{R},
\end{equation}
where $\MI(\cdot\,;\cdot)$ is mutual information and
$\mathcal{R}=\MI(d;s)+\MI(d;e)-\MI(d;s,e)$ is the redundancy, the information
about $d$ that $s$ duplicates (nonnegative in the redundant regime we
consider). A modality shortcut is the remainder $\MI(d;e\mid s)$ vanishing on
successful demonstrations, for one of two reasons. Under \emph{collinearity},
$\mathcal{R}\approx\MI(d;e)$: in \vcap{} the target fills the wrist view
because the operator reaches for it, and in \llock{} a noun always comes with
one verb. Under \emph{degeneracy}, the decision never varies,
$\Ent(d)\approx 0$: in \minert{} nothing follows gripper closure but a lift.
More demonstrations of the same kind preserve both, so scaling alone need not
remove the shortcut.

\Cref{eq:suffic} thus suggests two moves with different costs. Lowering
$\mathcal{R}$, as a decoy inserted off the grasp path does, makes the
shortcut insufficient on its own without changing the expert action, so the
labels come free. Raising $\MI(d;e)$, as captions naming decision-relevant
attributes do, makes the evidence easier to use, but under degeneracy
$\MI(d;e)\le\Ent(d)$ stays near zero until a missing branch, such as reopening
after an empty grasp, is recorded; relabeling cannot supply it. These are goals
for the training data, not a training loss or a guarantee of learned evidence
use.

\subsection{\method}
\label{sec:training}

\method{} trains on two sources: task demonstrations, which supervise
real-robot General PnP and, separately, the 50 RoboTwin~2.0 tasks
(\cref{sec:experiments}), and recorded random-motion and failed-execution
trajectories, screened and relabeled as auxiliary data rather than extra tasks.
Its three components build training examples from both while keeping every
action label valid (\cref{fig:overview}): V targets the redundancy
$\mathcal{R}$, C the evidence term $\MI(d;e)$, and R adds recorded segments in
which $d$ varies.

The visual component V applies task-preserving wrist-view perturbations $\TV$:
a plausible distractor of varying identity and placement is inserted off the
grasp path, without hiding the target or changing contact geometry, so local
salience changes while the task, motor context, and recorded action stay valid,
and the edit itself cannot become a task cue. An edit that would change the
target, collision constraints, or required grasp needs a new action and is
rejected. The intended lesson is that a salient object need not be the
instructed one, yet wrist evidence should still guide the motion; noise,
masking, and camera removal are controls, not substitutes.

The caption component C applies decision-relevant caption enrichment $\TC$
and leaves observations and actions unchanged. A vision--language model splits
each recording into atomic skills (grasp, lift, move, put down) and describes
each by the acting hand and whether it opens or closes, the operation, the
object with just enough attributes to single it out, and its position, initial
state, and grasp point (\cref{app:caption_prompt}): the grasp in \emph{pick up
the mug} becomes \emph{the left hand closes and grasps the upright blue mug
left of the tray by its body}. The aim is decision-relevant detail, not length;
a different verb appears only when the data contain that action, since
rewriting a grasp as a push does not create a push demonstration. Both sources
mix brief and detailed instructions under the same rules, so caption style does
not reveal the source, and the deployed policy takes ordinary instructions.

The trajectory component R applies caption-grounded trajectory expansion $\TR$
to recorded random motion and failed executions. These are divided into
segments with synchronized images, motor context, and actions; unusable
segments are discarded, and each retained one gets a brief instruction for what
it actually does, plus a detailed version from C. A failed attempt to pick up a
bowl may contain a valid segment \emph{move the empty gripper left and open
it}, whose label describes this local behavior, not the original goal and not a
failure that has yet to happen. R thus widens language-conditioned
state--action coverage but does not by itself teach recovery, which requires
recorded corrective continuations evaluated under the original instruction.

All three components train with the backbone's own action loss
$\mathcal L_{\mathrm{act}}$ (flow matching for $\pi_{0.5}$). With policy input
$\obs=(\vision,\instr,\prop)$, empirical distributions $\data_{\mathrm{demo}}$
of task demonstrations and $\data_{\mathrm{extra}}$ of relabeled auxiliary
segments, and an auxiliary fraction $\rho\in[0,1]$, we train on
\begin{equation}
\label{eq:objective}
\begin{aligned}
Q &=(1-\rho)\data_{\mathrm{demo}}+\rho\data_{\mathrm{extra}},\\
\mathcal L(\theta)
&=\mathbb E_{(\vision,\instr,\prop,\act)\sim Q}
\left[\mathcal L_{\mathrm{act}}
\bigl(\pol;\tilde\vision,\tilde\instr,\prop,\act\bigr)\right],
\end{aligned}
\end{equation}
where $\tilde\vision$ is the original or perturbed visual input, $\tilde\instr$
is a brief or detailed instruction for the same behavior, and the expectation
covers augmentation and caption sampling. Setting $\tilde\vision=\vision$,
$\tilde\instr=\instr$, or $\rho=0$ disables V, C, or R; R without C still uses
accurate brief instructions. All rates are fixed within a run and selected on
validation data. Because V, C, and R need not act only on the visual, lexical,
or motor shortcut family ($\chanV,\chanL,\chanA$), every configuration is
scored on all three (\cref{app:perturb_details,app:ablation}).

\subsection{\metric}
\label{sec:metric}

\begin{figure}[t]
\centering
\includegraphics[width=\linewidth]{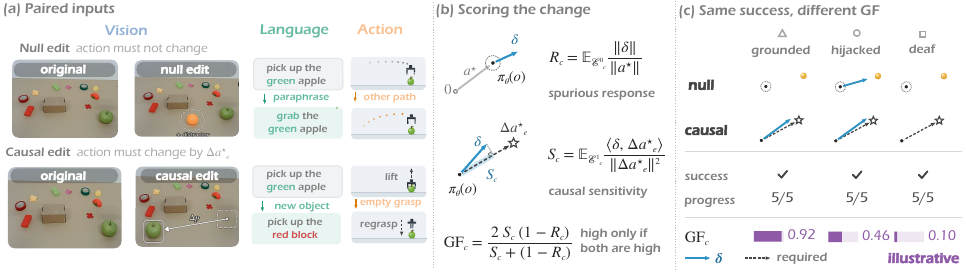}
\caption{\metric{} from paired inputs. (a) For each family, a null edit must
leave the action unchanged (an off-path decoy, a paraphrase, another approach to
the same pose), and a causal edit must change it by a known target
$\Delta \act^\star_e$ (the target moved by $\Delta p$, another object named, a
grasp that closed on nothing). (b) $R_\chan$ normalizes the predicted change
$\delta$; $S_\chan$ projects it onto $\Delta \act^\star_e$ and is clipped to
$[0,1]$. (c) Policies with equal success on the demonstration distribution can
differ sharply in \metric{}.}
\label{fig:groundfscore}
\end{figure}

Task success cannot separate a policy that reads the shortcut from one that
reads the evidence, since both succeed on the demonstration distribution.
\metric{} breaks this agreement with paired edits for each shortcut family
$\chan\in\{\chanV,\chanL,\chanA\}$ (\cref{fig:groundfscore}a): a \emph{null
edit} $e \in \Enull{\chan}$ alters the input without changing the correct
action, and a \emph{causal edit} $e \in \Ecaus{\chan}$ changes the correct
action by a known amount $\Delta \act^\star_e$. With $\act^\star$ the reference
action at the unedited observation $\obs$,
\begin{equation}
\label{eq:R}
R_\chan = \mathop{\mathbb{E}}_{\obs,\, e \sim \Enull{\chan}}
  \left[\frac{\lVert \pol(e(\obs)) - \pol(\obs) \rVert}{\lVert \act^\star \rVert}\right],
\qquad
S_\chan = \mathop{\mathbb{E}}_{\obs,\, e \sim \Ecaus{\chan}}
  \left[\frac{\langle \pol(e(\obs)) - \pol(\obs),\ \Delta \act^\star_e \rangle}
             {\lVert \Delta \act^\star_e \rVert^{2}}\right],
\end{equation}
where the \emph{spurious response} $R_\chan$ measures how much the policy moves
when it should not, and the \emph{causal sensitivity} $S_\chan$ is the fraction
of the demanded change that its response realizes ($1$ for exact compliance,
$0$ for none). Clipping both to $[0,1]$, we combine them as
\begin{equation}
\label{eq:gf}
\gf_\chan \;=\; \frac{2\, S_\chan\, (1 - R_\chan)}{S_\chan + (1 - R_\chan)},
\end{equation}
an F-score with precision $1 - R_\chan$ (not moving when nothing should move)
and recall $S_\chan$ (moving when something should). As in HalFscore
\citep{chen2025perturbollava}, the harmonic mean is high only when both terms
are: a policy that ignores the input has $R_\chan = 0$ but also $S_\chan = 0$,
and scores zero.

\metric{} needs only paired forward passes, not rollouts. We compute it on 100
held-out General PnP observations, each given null and causal edits of all
three families: language edits rewrite the instruction, visual edits are made
by image editing, and motor edits substitute recorded alternatives. For each
causal edit, $\Delta\act^\star_e$ is the difference between the action chunks
of two expert recordings from the same state, one under the original and one
under the edited condition. Actions are full predicted chunks in the normalized
action space of $\pi_{0.5}$, and the edited and unedited passes share the
flow-matching noise, so sampling noise does not count as a response;
$\lVert\act^\star\rVert$ is floored at a tenth of its median, and causal edits
with $\lVert\Delta\act^\star_e\rVert$ below a fifth of the median are
discarded. The score is a diagnostic, not a certificate of task competence;
\cref{sec:abl_metric} tests its association with online failures.

\section{Experiments}
\label{sec:experiments}

\subsection{Protocol}
In real-robot General PnP, the robot places the instructed object in the
instructed container. Pick-and-place underlies most longer tasks yet is far from
trivial here: each layout draws six objects and three
containers from 100 items, so combinations rarely repeat and evaluation layouts
are unseen in training. Distractors in every scene invite salience capture,
instructions require grounding the named object and container, and any grasp
can close on nothing, the setting of motor inertia. Short, uniform
demonstrations also make data scaling affordable (\cref{sec:abl_scaling});
RoboTwin~2.0 covers other skills (\cref{app:selected_tasks}).

\begin{figure}[t]
\centering
\includegraphics[width=\linewidth]{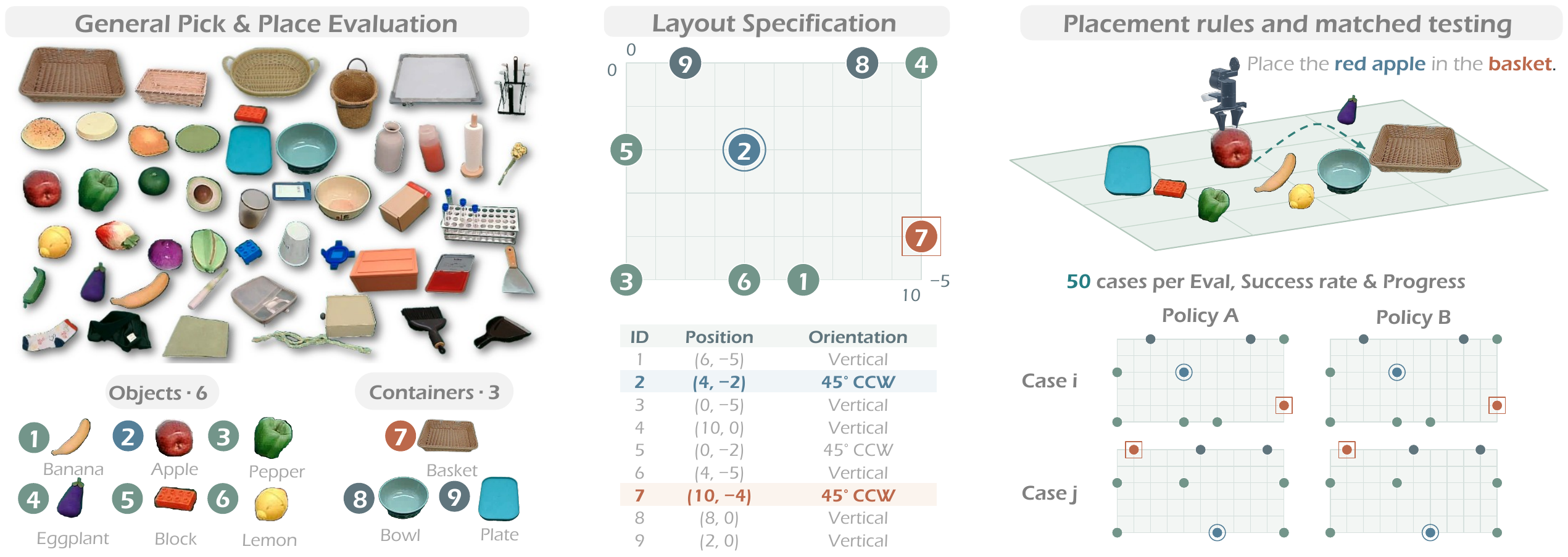}
\caption{Real-robot General PnP setup. Left: the material pool and the six
objects and three containers used in the task. Middle: an example stored
layout, specified by each item's position and orientation. Right: an
instruction-conditioned scene; every policy is tested on the same 50 stored
cases.}
\label{fig:task_coverage_ablations}
\end{figure}

The 50 evaluation layouts, generated once by rule-based random placement, are
restored before each rollout, so every policy runs the same cases once
(\cref{fig:task_coverage_ablations}, \cref{app:pnp_layouts}).

All experiments use $\pi_{0.5}$ \citep{black2025pi05}; conditions share
initialization, optimizer, training steps, and batch size, so each comparison
varies only the training data, with auxiliary examples replacing a fixed
fraction of each batch. Recordings and scene groups are split before
segmentation, captioning, or augmentation, validation selects the fixed rates,
and the primary result uses brief instructions (\cref{app:exp_details}).

\begin{table}[t]
\centering
\caption{Main results on real-robot General PnP with a $\pi_{0.5}$ backbone.
G1--G5 are success rates (\%) on five blocks of ten cases, SR is success over
all 50 cases, and Prog.\ is the mean 0--5 progress score. $\gf_V$, $\gf_L$, and
$\gf_A$ are \metric{} readouts on 100 held-out observations (\cref{sec:metric}); SC and MI are failure
rates (\%) on salience-capture and motor-inertia probes of 50 trials each,
separate from the 50 cases. $\pm$ is the half-width of the 95\% bootstrap
interval over the evaluation units: the ten cases of a block, the 50 cases for
SR and Prog., the 100 observations for \metric{}, and the 50 probe trials for SC
and MI. V, C, and R denote wrist-view perturbation, detailed captions, and
trajectory expansion; the upper block covers all eight combinations and the
lower block the controls. Bold marks the best value in each column; the shaded
row is the full method.}
\label{tab:main}
\setlength{\tabcolsep}{3.4pt}
\footnotesize
\resizebox{\linewidth}{!}{%
\begin{tabular}{@{}lccccccc@{\hspace{6pt}}ccc@{\hspace{6pt}}cc}
\toprule
& \multicolumn{7}{c}{Task success\,\up} & \multicolumn{3}{c}{\metric{}\,\up} & \multicolumn{2}{c}{Failure\,\down} \\
\cmidrule(lr){2-8}\cmidrule(lr){9-11}\cmidrule(lr){12-13}
Method & G1 & G2 & G3 & G4 & G5 & SR & Prog. & $\gf_V$ & $\gf_L$ & $\gf_A$ & SC & MI \\
\midrule
$\pi_{0.5}$ fine-tune    & \mn{40}\cim{30} & \mn{40}\cim{30} & \mn{50}\cim{30} & \mn{40}\cim{30} & \mn{40}\cim{30} & \mn{42}\cim{14} & \mn{2.71}\cim{0.55} & \mn{0.28}\cim{0.04} & \mn{0.24}\cim{0.03} & \mn{0.19}\cim{0.03} & \mn{68}\cim{13} & \mn{82}\cim{11} \\
+ V                      & \mn{50}\cim{30} & \mn{50}\cim{30} & \mn{60}\cim{30} & \mn{50}\cim{30} & \mn{50}\cim{30} & \mn{52}\cim{14} & \mn{3.12}\cim{0.54} & \mn{0.51}\cim{0.04} & \mn{0.26}\cim{0.03} & \mn{0.21}\cim{0.03} & \mn{36}\cim{13} & \mn{80}\cim{11} \\
+ C                      & \mn{60}\cim{30} & \mn{50}\cim{30} & \mn{50}\cim{30} & \mn{50}\cim{30} & \mn{60}\cim{30} & \mn{54}\cim{14} & \mn{3.20}\cim{0.53} & \mn{0.30}\cim{0.04} & \mn{0.47}\cim{0.04} & \mn{0.22}\cim{0.03} & \mn{62}\cim{14} & \mn{78}\cim{11} \\
+ R                      & \mn{50}\cim{30} & \mn{60}\cim{30} & \mn{50}\cim{30} & \mn{60}\cim{30} & \mn{60}\cim{30} & \mn{56}\cim{14} & \mn{3.34}\cim{0.52} & \mn{0.29}\cim{0.04} & \mn{0.27}\cim{0.03} & \mn{0.44}\cim{0.04} & \mn{64}\cim{13} & \mn{44}\cim{14} \\
+ V + C                  & \mn{70}\cim{30} & \mn{60}\cim{30} & \mn{60}\cim{30} & \mn{60}\cim{30} & \mn{70}\cim{30} & \mn{64}\cim{13} & \mn{3.62}\cim{0.50} & \mn{0.55}\cim{0.04} & \mn{0.50}\cim{0.04} & \mn{0.24}\cim{0.03} & \mn{30}\cim{13} & \mn{76}\cim{12} \\
+ V + R                  & \mn{60}\cim{30} & \mn{70}\cim{30} & \mn{70}\cim{30} & \mn{60}\cim{30} & \mn{70}\cim{30} & \mn{66}\cim{13} & \mn{3.73}\cim{0.48} & \mn{0.54}\cim{0.04} & \mn{0.28}\cim{0.04} & \mn{0.47}\cim{0.04} & \mn{34}\cim{13} & \mn{40}\cim{14} \\
+ C + R                  & \mn{70}\cim{30} & \mn{70}\cim{30} & \mn{60}\cim{30} & \mn{70}\cim{30} & \mn{70}\cim{30} & \mn{68}\cim{13} & \mn{3.82}\cim{0.47} & \mn{0.31}\cim{0.04} & \mn{0.51}\cim{0.04} & \mn{0.48}\cim{0.04} & \mn{58}\cim{14} & \mn{38}\cim{14} \\
\oursrow
\method{} (V + C + R)    & \mn{\textbf{90}}\cim{15} & \mn{\textbf{80}}\cim{25} & \mn{\textbf{80}}\cim{25} & \mn{\textbf{80}}\cim{25} & \mn{\textbf{90}}\cim{15} & \mn{\textbf{84}}\cim{10} & \mn{\textbf{4.36}}\cim{0.37} & \mn{\textbf{0.62}}\cim{0.04} & \mn{\textbf{0.57}}\cim{0.04} & \mn{\textbf{0.53}}\cim{0.04} & \mn{20}\cim{11} & \mn{\textbf{24}}\cim{12} \\
\midrule
Noise / dropout          & \mn{40}\cim{30} & \mn{50}\cim{30} & \mn{40}\cim{30} & \mn{40}\cim{30} & \mn{50}\cim{30} & \mn{44}\cim{14} & \mn{2.80}\cim{0.55} & \mn{0.33}\cim{0.04} & \mn{0.25}\cim{0.03} & \mn{0.20}\cim{0.03} & \mn{60}\cim{14} & \mn{82}\cim{11} \\
$2\times$ demonstrations & \mn{50}\cim{30} & \mn{40}\cim{30} & \mn{50}\cim{30} & \mn{50}\cim{30} & \mn{50}\cim{30} & \mn{48}\cim{14} & \mn{2.95}\cim{0.55} & \mn{0.29}\cim{0.04} & \mn{0.25}\cim{0.03} & \mn{0.20}\cim{0.03} & \mn{66}\cim{13} & \mn{80}\cim{11} \\
Negative guidance        & \mn{50}\cim{30} & \mn{40}\cim{30} & \mn{50}\cim{30} & \mn{40}\cim{30} & \mn{50}\cim{30} & \mn{46}\cim{14} & \mn{2.86}\cim{0.55} & \mn{0.41}\cim{0.04} & \mn{0.36}\cim{0.04} & \mn{0.28}\cim{0.04} & \mn{48}\cim{14} & \mn{62}\cim{14} \\
Wrist camera removed     & \mn{30}\cim{30} & \mn{20}\cim{25} & \mn{30}\cim{30} & \mn{20}\cim{25} & \mn{30}\cim{30} & \mn{26}\cim{12} & \mn{1.94}\cim{0.54} & \mn{0.16}\cim{0.03} & \mn{0.30}\cim{0.04} & \mn{0.18}\cim{0.03} & \mn{\textbf{12}}\cim{9} & \mn{88}\cim{9} \\
\bottomrule
\end{tabular}}
\end{table}

\subsection{General Pick and Place}
\cref{tab:main} covers all eight combinations of V, C, and R, so each component
appears alone and as a leave-one-out. An off-path wrist distractor probes
salience capture (SC) and a grasp scripted to close on nothing probes motor
inertia (MI); verb-change probes for noun lock-in run in RoboTwin~2.0
(\cref{app:verb_probes}). The controls test generic
regularization (noise and dropout at the V budget), data quantity (twice the
demonstrations), discarding evidence (camera removal), and inference-time
correction (negative guidance; \cref{app:exp_details}).

Each component raises success and mainly the \metric{} of its target family: V
lifts $\gf_V$ from \mn{0.28} to \mn{0.51} and cuts SC from \mn{68}\% to
\mn{36}\%, C lifts $\gf_L$ from \mn{0.24} to \mn{0.47}, and R lifts $\gf_A$
from \mn{0.19} to \mn{0.44} and cuts MI from \mn{82}\% to \mn{44}\%. The gains
compound: every pair beats both of its members, and the full method doubles
baseline success to \mn{84}\% with the best readout in all three families,
while removing any one component costs \mn{16}--\mn{20} points. The other
controls stay within \mn{6} points of the baseline, and camera removal attains
the lowest SC only by losing \mn{16} points of success and lowering $\gf_V$ to
\mn{0.16}.

\subsection{Generality across skills}
RoboTwin~2.0 \citep{chen2025robotwin2} tests whether \method{} generalizes
beyond General PnP. Its 50 bimanual tasks span skills such as handing over a
block, opening a microwave, hammering, and hanging a mug, and its domain
randomization varies textures, lighting, clutter, tabletop height, and
instruction wording. We follow the two settings of \citet{cai2026internw0}:
Full trains on clean and randomized demonstrations, whereas Clean2Random trains
only on clean ones, so randomized scenes are unseen; both evaluate on clean and
randomized scenes. Clean2Random is the more direct test of modality shortcuts,
since randomized clutter adds salient cues that clean training never
contradicted. R uses simulator rollouts relabeled as in the real-robot study
(\cref{app:selected_tasks}), and all variants share one budget; published
results, from different architectures and recipes, give context only.

\method{} improves the $\pi_{0.5}$ baseline in every column of
\cref{tab:task_coverage}, most where shortcuts should matter: on randomized
scenes in Clean2Random it gains \mn{12.2} points, against \mn{2.7} on clean
scenes, the best randomized-scene result among the listed systems. Removing V
costs most on randomized scenes in both settings (\mn{8.4} and \mn{2.6}
points), consistent with its role against salient distractors.

\begin{table}[t]
\centering
\caption{RoboTwin~2.0 results (task-averaged success, \%). Avg.\ is the mean
of clean and randomized scenes. Published results are as reported by
\citet{cai2026internw0}, rounded to one decimal, and -- marks an unreported
setting; \cref{tab:task_coverage_full} lists every published result, including
systems with higher reported scores. The lower block uses our $\pi_{0.5}$
fine-tuning recipe. Bold marks the best value in each column; the shaded row is
the full method.}
\label{tab:task_coverage}
\footnotesize
\setlength{\tabcolsep}{5pt}
\begin{tabular}{@{}lccc@{\hspace{10pt}}ccc}
\toprule
& \multicolumn{3}{c}{Full} & \multicolumn{3}{c}{Clean2Random} \\
\cmidrule(lr){2-4}\cmidrule(lr){5-7}
Method & Clean & Rand. & Avg. & Clean & Rand. & Avg. \\
\midrule
$\pi_0$ \citep{black2025pi0}                   & 65.9 & 58.4 & 62.2 & -- & -- & -- \\
$\pi_{0.5}$ \citep{black2025pi05}              & 82.7 & 76.8 & 79.8 & 70.7 & 46.0 & 58.4 \\
ABot-M0 \citep{yang2026abotm0}                 & 81.2 & 80.4 & 80.8 & -- & -- & -- \\
X-VLA \citep{zheng2025xvla}                    & -- & -- & -- & 68.0 & 20.9 & 44.5 \\
X-WAM \citep{guo2026xwam}                      & -- & -- & -- & 70.0 & 25.8 & 47.9 \\
Spatial Forcing \citep{li2025spatialforcing}   & -- & -- & -- & 77.2 & 26.7 & 52.0 \\
4D-WAM \citep{yang2026fourdwam}                & -- & -- & -- & \textbf{81.5} & 41.8 & 61.6 \\
\midrule
\method{} w/o V       & \mn{84.9} & \mn{79.3} & \mn{82.1} & \mn{72.6} & \mn{49.8} & \mn{61.2} \\
\method{} w/o C       & \mn{84.1} & \mn{80.6} & \mn{82.4} & \mn{71.8} & \mn{55.4} & \mn{63.6} \\
\method{} w/o R       & \mn{84.4} & \mn{80.9} & \mn{82.7} & \mn{72.0} & \mn{56.2} & \mn{64.1} \\
\oursrow
\method{}             & \mn{\textbf{85.6}} & \mn{\textbf{81.9}} & \mn{\textbf{83.8}} & \mn{73.4} & \mn{\textbf{58.2}} & \mn{\textbf{65.8}} \\
\bottomrule
\end{tabular}
\end{table}

\section{Ablation study}
\label{sec:ablation}

\begin{figure}[t]
\centering
\includegraphics[width=\linewidth]{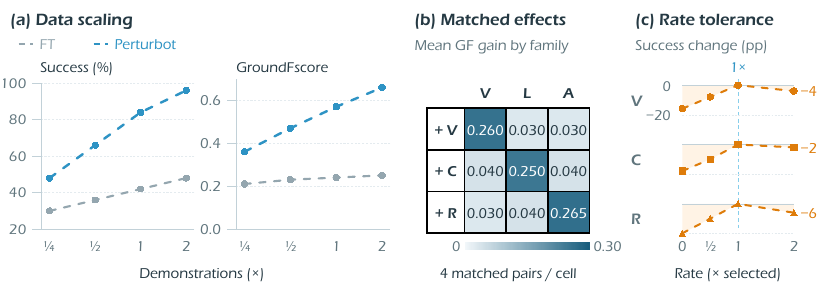}
\caption{Scaling and ablation effects with a $\pi_{0.5}$ backbone. (a) Success
(left) and mean \metric{} (right) on a log-scaled demonstration budget;
\cref{tab:scaling} compares their slopes. (b) Mean \metric{} gain per family
from adding V, C, or R, averaged over four contrasts with the other components
fixed (\cref{tab:main}). (c) Change in success from the selected rate as one
sampling rate varies (\cref{tab:sensitivity}); shading shows the loss, not
uncertainty.}
\label{fig:ablation_scaling}
\end{figure}

We test whether \method{}'s gains require new rather than more data, are
mechanism-specific, and preserve control, and whether \metric{} predicts online
failures. Unless noted, ablations use real-robot
General PnP with $\pi_{0.5}$ (\cref{fig:ablation_scaling}; full tables in
\cref{app:abl_extra}).

\subsection{Data scaling}
\label{sec:abl_scaling}
The default remedy for a brittle policy is more demonstrations, yet
\cref{sec:scaling} predicts that same-kind ones preserve the collinearity and
degeneracy behind shortcuts, so scaling tests whether the data must change.
We train the $\pi_{0.5}$ baseline and \method{} on nested
subsets of 125, 250, 500 ($1\times$), and all 1{,}000 General PnP
demonstrations at fixed optimizer steps, three seeds each, and report each
curve's least-squares slope per doubling (\cref{tab:scaling}).

\begin{table}[t]
\centering
\begin{minipage}[t]{0.51\linewidth}
\centering
\caption{Data scaling at fixed optimizer steps. $\overline{\gf}$: mean
\metric{}; slope: change per doubling; $\rho$: SR--$\overline{\gf}$
correlation over 4 budgets $\times$ 3 seeds.}
\label{tab:scaling}
\setlength{\tabcolsep}{3pt}
\renewcommand{\arraystretch}{1.51}
\resizebox{\linewidth}{!}{%
\begin{tabular}{lcccccccc}
\toprule
& \multicolumn{4}{c}{FT} & \multicolumn{4}{c}{\method{}} \\
\cmidrule(lr){2-5}\cmidrule(lr){6-9}
Demos & SR\,\up & $\overline{\gf}$\,\up & SC\,\down & MI\,\down & SR\,\up & $\overline{\gf}$\,\up & SC\,\down & MI\,\down \\
\midrule
$1/4\times$ & \mn{30} & \mn{0.21} & \mn{74} & \mn{86} & \mn{48} & \mn{0.36} & \mn{38} & \mn{46} \\
$1/2\times$ & \mn{36} & \mn{0.23} & \mn{70} & \mn{84} & \mn{66} & \mn{0.47} & \mn{28} & \mn{34} \\
$1\times$   & \mn{42} & \mn{0.24} & \mn{68} & \mn{82} & \mn{84} & \mn{0.57} & \mn{20} & \mn{24} \\
$2\times$   & \mn{48} & \mn{0.25} & \mn{66} & \mn{80} & \mn{96} & \mn{0.66} & \mn{14} & \mn{18} \\
\midrule
Slope  & \mn{+6.0} & \mn{+0.013} & \mn{$-$2.6} & \mn{$-$2.0} & \mn{\textbf{+16.2}} & \mn{\textbf{+0.100}} & \mn{\textbf{$-$8.0}} & \mn{\textbf{$-$9.4}} \\
$\rho$ & \multicolumn{2}{c}{\mn{0.41}} & & & \multicolumn{2}{c}{\mn{\textbf{0.93}}} & & \\
\bottomrule
\end{tabular}}
\end{minipage}\hfill
\begin{minipage}[t]{0.465\linewidth}
\centering
\caption{Budget-matched controls. Each control matches the budget of the
component below it but removes its mechanism.}
\label{tab:mechanism}
\setlength{\tabcolsep}{3pt}
\resizebox{\linewidth}{!}{%
\begin{tabular}{lcccccc}
\toprule
Condition & SR\,\up & $\gf_V$\,\up & $\gf_L$\,\up & $\gf_A$\,\up & SC\,\down & MI\,\down \\
\midrule
FT             & \mn{42} & \mn{0.28} & \mn{0.24} & \mn{0.19} & \mn{68} & \mn{82} \\
\midrule
Generic aug.   & \mn{44} & \mn{0.33} & \mn{0.25} & \mn{0.20} & \mn{60} & \mn{82} \\
Random decoys  & \mn{40} & \mn{0.38} & \mn{0.24} & \mn{0.19} & \mn{50} & \mn{82} \\
+ V            & \mn{52} & \mn{\textbf{0.51}} & \mn{0.26} & \mn{0.21} & \mn{\textbf{36}} & \mn{80} \\
\midrule
Paraphrases    & \mn{44} & \mn{0.28} & \mn{0.28} & \mn{0.19} & \mn{66} & \mn{82} \\
Length-matched & \mn{44} & \mn{0.29} & \mn{0.30} & \mn{0.20} & \mn{66} & \mn{80} \\
+ C            & \mn{54} & \mn{0.30} & \mn{\textbf{0.47}} & \mn{0.22} & \mn{62} & \mn{78} \\
\midrule
Success demos  & \mn{46} & \mn{0.29} & \mn{0.25} & \mn{0.20} & \mn{66} & \mn{80} \\
Random only    & \mn{50} & \mn{0.29} & \mn{0.26} & \mn{0.34} & \mn{66} & \mn{60} \\
Failed only    & \mn{52} & \mn{0.29} & \mn{0.26} & \mn{0.39} & \mn{64} & \mn{52} \\
+ R            & \mn{56} & \mn{0.29} & \mn{0.27} & \mn{\textbf{0.44}} & \mn{64} & \mn{\textbf{44}} \\
\bottomrule
\end{tabular}}
\end{minipage}
\end{table}

\method{} scales faster, gaining \mn{16.2} points of success per doubling against
\mn{6.0} for fine-tuning; at $1/4\times$ it already matches fine-tuning at
$2\times$, with eight times fewer demonstrations. The gains also differ in kind: from $1/4\times$ to $2\times$, fine-tuning adds
\mn{18} points while $\overline{\gf}$ moves by only \mn{0.04} and both failure
rates stay high; extra demonstrations add proficiency but not reliance on task
evidence, the unhealthy regime \metric{} is meant to expose. Under \method{},
success and $\overline{\gf}$ rise together, by \mn{48} points and \mn{0.30},
and both failure rates fall at every budget. Success and $\overline{\gf}$
correlate at Spearman $\rho=\mn{0.93}$ for \method{} but \mn{0.41} for
fine-tuning, whose $\overline{\gf}$ barely varies, so seed noise dominates its
ranking.

\subsection{Component specificity}
\label{sec:abl_mechanism}
Every component adds training data, so its gain could come from more samples or
generic regularization rather than from the shortcut it targets
(\cref{eq:suffic}). Each is therefore compared with controls that match its
budget but remove its mechanism (\cref{tab:mechanism}): for V, noise, cropping,
and dropout, or the same distractors placed anywhere in the wrist view,
including on the grasp path; for C, paraphrases or length-matched text that
does not bear on the decision; for R, equal-duration successful demonstrations
or either auxiliary source alone. Only the mechanism moves the targeted
readout. V reaches a $\gf_V$ of \mn{0.51} against at most \mn{0.38}, and random
placement drops success below the baseline, as on-path decoys invalidate the
recorded action. Paraphrases and length-matched text raise $\gf_L$ by at most
\mn{0.06}, against \mn{0.23} for C. Extra successful demonstrations leave MI
near the baseline, whereas random and failed segments each lower it and
together reach \mn{44}\%.

\subsection{Controllability}
\label{sec:abl_control}
A policy can pass shortcut probes by ignoring an input, as removing the wrist
camera does for salience capture (\cref{tab:main}), but such robustness
discards evidence the task needs. \cref{tab:control} therefore pairs changes
that should not alter the action with changes that should: a distractor should
not redirect a reach, yet the same object should be selected when it is
instructed, and lifting should depend on the observed grasp outcome.
\method{} picks the distractor in \mn{20}\% against \mn{68}\% for fine-tuning,
yet still selects the instructed object in \mn{90}\%; without a wrist camera, a
policy avoids the distractor more often but finds the instructed object in only
\mn{48}\%. \method{} also lifts secured objects nearly
as often as fine-tuning (\mn{92}\% vs.\ \mn{94}\%) but lifts after empty grasps
in \mn{24}\% rather than \mn{82}\%.

\begin{table}[t]
\centering
\begin{minipage}[t]{0.63\linewidth}
\centering
\caption{Controllability on real-robot General PnP (\%). Target-selection
probes change which object is instructed; grasp-outcome probes compare a
secured object with an empty grasp. Each probe has 50 trials per policy,
separate from the 50 cases.}
\label{tab:control}
\footnotesize
\setlength{\tabcolsep}{2.4pt}
\begin{tabular}{lcccc}
\toprule
& \multicolumn{2}{c}{Target selected} & \multicolumn{2}{c}{Lift executed} \\
\cmidrule(lr){2-3}\cmidrule(lr){4-5}
Condition & instructed\,\up & distractor\,\down & secured\,\up & empty\,\down \\
\midrule
$\pi_{0.5}$ fine-tune & \mn{88} & \mn{68} & \mn{94} & \mn{82} \\
No wrist camera       & \mn{48} & \mn{12} & \mn{90} & \mn{88} \\
\method{}             & \mn{\textbf{90}} & \mn{20} & \mn{92} & \mn{\textbf{24}} \\
\bottomrule
\end{tabular}
\end{minipage}\hfill
\begin{minipage}[t]{0.33\linewidth}
\centering
\caption{Spearman $\rho$ with the negative online failure rate over 36
checkpoints; intervals in \cref{tab:gf_ci}. $^\dagger$Ranks the no-wrist
policy above the baseline.}
\label{tab:gf_terms}
\footnotesize
\setlength{\tabcolsep}{4pt}
\begin{tabular}{lc}
\toprule
Predictor & $\rho$\,\up \\
\midrule
Task success                 & \mn{0.31} \\
Stability ($1-R$)$^\dagger$  & \mn{0.44} \\
Responsiveness ($S$)         & \mn{0.38} \\
\metric{}                    & \mn{\textbf{0.72}} \\
\bottomrule
\end{tabular}
\end{minipage}
\end{table}

\subsection{Validity of \metric{}}
\label{sec:abl_metric}
\metric{} is meant to replace costly shortcut-targeted rollouts with offline
forward passes, which is only useful if it predicts what those rollouts reveal.
Across the twelve conditions in \cref{tab:main} and three seeds, we
rank-correlate \metric{} on the 100 paired observations of \cref{sec:metric}
with the negative online failure rate on shortcut-targeted trials, and ablate
the score itself: stability alone ($1-R$) rewards ignoring an input, and
responsiveness alone ($S$) rewards moving under any change. \metric{} predicts
online failures best ($\rho=\mn{0.72}$), and its advantage of \mn{0.41} over
task success has a 95\% cluster-bootstrap interval of
[\mn{0.12}, \mn{0.68}] (\cref{tab:gf_ci}). Neither term suffices alone:
stability reaches \mn{0.44} but ranks the no-wrist policy above the baseline,
and responsiveness reaches only \mn{0.38}.

\section{Limitations}
\label{sec:limitations}

\method{} needs detailed captions and relabeled segments annotated to match the
recorded behavior, and random and failed trajectories in addition to
demonstrations; relabeling cannot create behaviors that were never recorded,
such as recovery. Real-robot rollouts offer a ready source of such trajectories:
successful trajectories tend to resemble one another, while each failure goes
wrong in its own way.

\section{Conclusion}
\label{sec:conclusion}

VLA policies can succeed on familiar tasks while relying on modality shortcuts:
visual, lexical, or motor cues that successful demonstrations make sufficient to
predict the expert's action, and that more demonstrations of the same kind keep
predictive. \method{} changes what the training data demand through
task-preserving wrist-view perturbations, decision-relevant captions, and
relabeled random and failed segments, without changing inference, and \metric{}
complements task success by diagnosing, offline, whether a policy improves
through task evidence or through shortcuts. Together, they point to a different
aim for scaling robot data: expanding not only repetitions of familiar behavior
but also the range of decisions that depend on the scene, instruction, and
observed outcome.

\label{end:maintext}

\subsection*{AI use statement}

In this work, we used generative AI tools to polish the writing for
readability, including rephrasing sentences and reorganizing sentence order and
logical flow. All AI-assisted text was reviewed and verified by the authors, who
take full responsibility for the final content of this work, including all
text, claims, and artifacts.

\subsection*{Reproducibility statement}

We will open-source the data, the trained checkpoints, and the models upon
publication. The main modeling relation is \cref{eq:suffic}, and the \metric{}
equations and probes are specified in \cref{sec:metric}. \cref{app:exp_details} describes the real-robot
platform and evaluation protocols, \cref{app:perturb_details} the training data
and its collection, \cref{app:training} the $\pi_{0.5}$ configuration,
\cref{app:wrist_perturbation} the wrist-view perturbation, and
\cref{app:caption_prompt} the prompt used to generate detailed captions; the
common action-training objective is \cref{eq:objective}.

\clearpage
\bibliographystyle{iclr2027_conference}
\bibliography{perturbot}

\appendix
\clearpage
\section*{Appendix}

\section{Experimental setup}
\label{app:exp_details}

\subsection{Real-robot platform}
\label{app:robot_setup}

\begin{figure}[ht]
\centering
\begin{minipage}[c]{0.46\linewidth}
\centering
\includegraphics[width=0.5\linewidth]{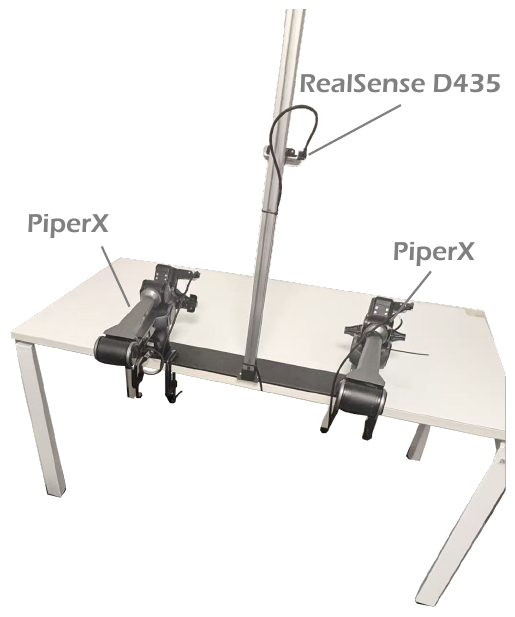}
\end{minipage}\hfill
\begin{minipage}[c]{0.5\linewidth}
\small
\captionof{figure}{Real-robot platform. Two AgileX PiPER~X arms are mounted
side by side and face the tabletop workspace. A \mbox{RealSense D435} head camera on
the central vertical post provides the third-person view, and a RealSense wrist
camera beside each gripper provides a wrist view.}
\label{fig:robot_setup}
\end{minipage}
\end{figure}

All real-robot data (task demonstrations, random-motion and failed-execution
recordings, and evaluation rollouts) are collected on the dual-arm AgileX
PiPER~X platform in \cref{fig:robot_setup}. Two 6-DoF arms, each with a
two-finger parallel gripper, are mounted side by side facing the tabletop. A
\mbox{RealSense D435} head camera is mounted on the central vertical post, and a
RealSense wrist camera is mounted on each arm. All three streams are recorded
at 30\,FPS and $640\times360$ pixels. The head camera supplies the
third-person view $\third$ and the wrist cameras supply $\wrist$; the three
views are resized to the policy input resolution (\cref{tab:hparams}) and form
the visual input $\vision$. The motor context $\prop$ is the proprioceptive
state of both arms, and the action $\act$ commands both arms and their grippers.
Demonstrations and auxiliary recordings are collected by teleoperation.

\subsection{General PnP evaluation}
\label{app:pnp_layouts}

The real-robot study evaluates only General PnP: six objects and three
containers on the tabletop, with instructions such as \emph{place the red apple
in the basket} (\cref{fig:task_coverage_ablations}). Placement rules are fixed
before evaluation. Each of the 50 cases specifies the object identities, target,
destination, and instruction; object and container positions are drawn at random
in a common tabletop frame, and each object's orientation follows its placement
rule. Conflicts are resolved by fixed rules (a distractor overlapping the target
is moved rather than the target, and an object--container conflict moves the
container), and the adjusted coordinates are stored. The 50 layouts are
generated once and restored before every rollout, so all policies and training
seeds face the same cases.

Each checkpoint executes every case once. We report success over the 50 cases,
the success of the five blocks G1--G5 (cases 1--10, \dots, 41--50), and the mean
0--5 progress score. The primary evaluation uses brief instructions; a second
pass replays the same 50 cases with detailed instructions
(\cref{app:detailed_eval}). The salience-capture, motor-inertia,
target-selection, and grasp-outcome probes use separate trial lists and add no
physical-robot task; each list has 50 trials per policy. The intervals in \cref{tab:main} are 95\% percentile bootstrap
intervals from 10{,}000 resamples of the evaluation units of one checkpoint
(cases, probe trials, or \metric{} observations), reported as half-widths.

\subsection{RoboTwin~2.0}
\label{app:selected_tasks}

RoboTwin~2.0 is evaluated on its 50-task suite in the two settings of
\citet{cai2026internw0}. Full trains on the official \texttt{demo\_clean} and
\texttt{demo\_randomized} demonstrations, whereas Clean2Random trains only on
\texttt{demo\_clean}; both evaluate on held-out clean and randomized trials, so
Clean2Random tests unseen randomized scenes. Within a setting, all methods share
the simulator randomization, demonstration pool, initial-state seeds, and trial
budget, and the simulator's domain randomization is not counted as V.
Verb-change probes for noun lock-in use only task pairs that admit both
operations from matched states with valid demonstrations
(\cref{app:verb_probes}). \cref{tab:task_coverage_full} lists all published results reported by
\citet{cai2026internw0}.

\paragraph{Auxiliary trajectories.}
In simulation, R uses rollouts instead of teleoperated recordings.
Random-motion trajectories start from a task's initial state and execute
scripted, randomly sampled arm motions that sweep the workspace, as in the
real-robot protocol. Failed trajectories come from the simulator's scripted
expert with injected perturbations that make a grasp or placement go wrong,
such as offset grasp poses, early gripper release, or shifted placement
targets; only rollouts that the success check marks as failed are kept, so they
contain the near misses, empty grasps, and misplacements that successful
demonstrations lack. Because the simulator
records exact states and actions, both kinds are segmented, captioned, and
relabeled with the same rules as the real recordings
(\cref{app:perturb_details}). In Clean2Random, rollouts use clean scenes only,
so no randomized scene enters training.

\begin{table}[t]
\centering
\caption{All published RoboTwin~2.0 results reported by
\citet{cai2026internw0} (task-averaged success, \%), rounded to one decimal;
-- marks an unreported setting. They come from different architectures and
training recipes; \cref{tab:task_coverage} gives the matched $\pi_{0.5}$
comparison.}
\label{tab:task_coverage_full}
\small
\setlength{\tabcolsep}{5pt}
\begin{tabular}{lcccccc}
\toprule
& \multicolumn{3}{c}{Full} & \multicolumn{3}{c}{Clean2Random} \\
\cmidrule(lr){2-4}\cmidrule(lr){5-7}
Method & Clean & Rand. & Avg. & Clean & Rand. & Avg. \\
\midrule
$\pi_0$ \citep{black2025pi0}                   & 65.9 & 58.4 & 62.2 & -- & -- & -- \\
$\pi_{0.5}$ \citep{black2025pi05}              & 82.7 & 76.8 & 79.8 & 70.7 & 46.0 & 58.4 \\
ABot-M0 \citep{yang2026abotm0}                 & 81.2 & 80.4 & 80.8 & -- & -- & -- \\
Motus \citep{bi2025motus}                      & 88.7 & 87.0 & 87.8 & -- & -- & -- \\
Fast-WAM \citep{yuan2026fastwam}               & 91.9 & 91.8 & 91.8 & 77.8 & 1.9 & 39.9 \\
LingBot-VA \citep{li2026lingbotva}             & 92.9 & 91.5 & 92.2 & -- & -- & -- \\
AHA-WAM \citep{cai2026ahawam}                  & 93.4 & 92.2 & 92.8 & -- & -- & -- \\
OpenWAM-$\alpha$ \citep{wang2026openwam}       & 93.7 & 93.5 & 93.6 & -- & -- & -- \\
ABot-M0.5 \citep{chen2026abotm05}              & 94.0 & 94.2 & 94.1 & -- & -- & -- \\
X-VLA \citep{zheng2025xvla}                    & -- & -- & -- & 68.0 & 20.9 & 44.5 \\
X-WAM \citep{guo2026xwam}                      & -- & -- & -- & 70.0 & 25.8 & 47.9 \\
Spatial Forcing \citep{li2025spatialforcing}   & -- & -- & -- & 77.2 & 26.7 & 52.0 \\
4D-WAM \citep{yang2026fourdwam}                & -- & -- & -- & 81.5 & 41.8 & 61.6 \\
GigaBrain-0.7 \citep{gigabrain2026gigabrain07} & -- & -- & -- & 66.8 & 67.9 & 67.3 \\
InternW0 \citep{cai2026internw0}               & 93.2 & 93.0 & 93.1 & 83.2 & 68.0 & 75.6 \\
\bottomrule
\end{tabular}
\end{table}



\section{Training data}
\label{app:perturb_details}

\cref{tab:data_inventory} summarizes the real-robot training data. All of it
is collected on the General PnP setup: task demonstrations supervise the task,
and the auxiliary recordings supply R, never complete the task, and are not
evaluation cases.

\begin{table}[t]
\centering
\caption{Real-robot training data, all collected on the General PnP setup.
Main experiments use a fixed subset of 500 demonstrations ($1\times$) and all
auxiliary recordings; the $2\times$ budget uses all 1{,}000 demonstrations.}
\label{tab:data_inventory}
\small
\setlength{\tabcolsep}{6pt}
\begin{tabular}{llr}
\toprule
Source & Collection & Recordings \\
\midrule
General PnP demonstrations & teleoperated task executions & 1{,}000 \\
Random motion & free arm motion among the objects & 200 \\
Failed execution & deliberate contact attempts and failures & 200 \\
\midrule
Total & & 1{,}400 \\
\bottomrule
\end{tabular}
\end{table}

\subsection{General PnP demonstrations}

We collect 1{,}000 teleoperated demonstrations of the task, each with its brief
instruction. All main conditions train on a fixed subset of 500 ($1\times$); the
$1/4\times$ and $1/2\times$ budgets are nested subsets of it, the $2\times$
budget and the $2\times$ demonstrations control use all 1{,}000, and the
equal-duration successful demonstrations of the R control come from the 500
outside the $1\times$ subset. Items come from a pool of 100 objects and
containers (\cref{fig:task_coverage_ablations}); each layout places nine of them,
six objects and three containers as in evaluation, at random positions, and the
50 evaluation layouts are never used for training. Recordings and scene groups
are split into training, validation, and evaluation before any segmentation,
captioning, or augmentation, and all derivatives of a recording stay in its
split.

\subsection{Random-motion and failed-execution recordings}

The auxiliary recordings are collected to produce varied contacts and contact
failures rather than completed tasks: the arm approaches objects, touches them,
and occasionally touches the wrong one, pushes in the wrong direction, or grasps
nothing. Each recording lasts 5--10\,s and contains three to six distinct
motions, without long pauses and without completing a task. Each uses a layout
of nine items from the same pool as the demonstrations, with a randomly varied
tablecloth color.

\paragraph{Random motion (200 recordings).}
With the items of a layout randomly placed, the operator moves the arm through
three to six consecutive motions (left--right, forward--backward, up--down,
wrist rotation, and gripper opening or closing), covering the workspace without
idling, reaching its edge, or repeating one motion, and then resets. Each
recording must show clear motion with changes of direction and height. These
recordings expose how commanded actions move the robot.

\paragraph{Failed execution (200 recordings).}
Each recording brings the gripper within 2--5\,cm above or 1--3\,cm beside an
object, makes one contact attempt (pressing down, pushing horizontally, or
closing on part of the object), follows with one variation
(pushing again, changing direction, lifting slightly and setting down, or
rotating the wrist), and leaves. Operators deliberately make small errors: near
misses, off-center or wrong-direction pushes, empty grasps, grasps that slip or
are released after a short lift, and pushes that barely move the object. They
act quickly, as in play rather than task execution. The recordings are balanced
across three outcomes: a deliberate miss, a light touch, and a clear
interaction, for example tipping a bowl. They expose the consequences of
contact, including the empty grasps and failed pushes that successful
demonstrations never show.

\subsection{Segmentation, relabeling, and mixture}

Each auxiliary recording is assigned to a split and then divided into
atomic-skill segments with synchronized images, motor context, and actions
(\cref{app:captions}); unusable segments are discarded. Each retained segment
receives a correct brief instruction for its local behavior and a detailed
caption. The original goal of a failed attempt is never kept as its label, an
action chunk must stay inside the segment its instruction describes, and
measured states and recorded actions are never overwritten. Because the
recordings contain no corrective continuations, R widens state--action coverage
but does not demonstrate recovery under the original goal.

Training samples task demonstrations with probability $1-\rho$ and auxiliary
segments with probability $\rho$ (\cref{eq:objective}); the wrist-perturbation
probability $p_V$ and the detailed-caption probability $p_C$ apply to both
sources, so caption style does not reveal the source. The three rates are fixed
within a run and selected on validation data. Disabling V, C, or R sets $p_V$,
$p_C$, or $\rho$ to zero, and R without C still uses correct brief
instructions. Auxiliary segments replace a fraction of each batch rather than
adding optimizer updates.

\section{\texorpdfstring{$\pi_{0.5}$}{pi0.5} training}
\label{app:training}

We fine-tune $\pi_{0.5}$ \citep{black2025pi05} with the configuration in
\cref{tab:hparams}, shared by all conditions; only the data construction
varies. The policy receives the head and both wrist views, the instruction, and
the dual-arm proprioceptive state, and is trained with its flow-matching action
loss to predict chunks of $H=16$ actions, about one second at the 15\,Hz control
rate. All modules are fine-tuned. Every condition and data-scaling budget uses
the same optimizer steps, each run takes about four days on eight GPUs, and the
enabled values of $p_V$, $p_C$, and $\rho$ are shared across the factorial
comparisons.

\begin{table}[t]
\centering
\caption{Fine-tuning configuration for $\pi_{0.5}$, shared by all conditions.}
\label{tab:hparams}
\small
\setlength{\tabcolsep}{6pt}
\begin{tabular}{ll}
\toprule
Setting & Value \\
\midrule
Adaptation              & full fine-tuning \\
Optimizer               & AdamW, $(\beta_1,\beta_2,\epsilon)=(0.9,\,0.999,\,10^{-8})$ \\
Learning rate           & $2\times10^{-5}$, constant, no warmup \\
Weight decay            & 0 \\
Gradient clipping       & 1.0 \\
Batch size              & 256 (32 per GPU, no accumulation) \\
Precision               & BF16 mixed precision \\
Image input             & $224\times224$ per view \\
Chunk length $H$        & 16 \\
Flow matching           & 4 noise samples per example; 10 Euler steps at inference \\
Hardware                & $8\times$ NVIDIA A100 80\,GB \\
Training time per run   & about 4 days \\
\bottomrule
\end{tabular}
\end{table}

\section{Wrist-view perturbation}
\label{app:wrist_perturbation}

V edits wrist-view images with GPT-Image-2.5 (\texttt{gpt-image-2.5-flare}),\footnote{OpenAI,
\url{https://openai.com/index/introducing-chatgpt-images-2-5/}.} an image
generation and editing model, to add plausible distractor objects. An edit is
kept only if the instructed target, the relevant spatial relations, the grasp
path, and the contact evidence remain readable and no new collision is implied;
otherwise it is rejected rather than given the original label. Only the wrist
image changes: the head view, motor state, instruction, and action label stay as
recorded. Distractor identity, appearance, and placement vary independently of
the task, and the training distractors are disjoint from the decoys used by the \metric{}
probes (\cref{sec:metric}). Generic corruption (noise,
cropping, and dropout at the same rate) and wrist-camera removal are separate
controls, and the salience-capture probes use physical distractors, so a policy
cannot pass them by detecting editing artifacts.

\section{Detailed-caption annotation}
\label{app:captions}

Each behavior has a brief instruction and a detailed caption. Detailed captions
are generated offline by a vision--language model with the prompt in
\cref{app:caption_prompt}: the model receives the timestamped frames of an
entire recording, splits it into contiguous atomic-skill segments, and describes
each segment by the acting hand and whether it opens or closes, the operation,
the position, the object's state, its distinguishing attributes, the object, and
the grasp point. Attributes such as color or position are added only when needed
to single out the object, and duplicates are distinguished by counting from one
side, so captions carry decision-relevant detail rather than length. Each
description uses only features visible in the frames and is checked against the
images and action record; uncertain object identities, motion directions, or
contact descriptions are corrected or excluded. A different verb is used only
when the recorded action matches it, and the same rules apply to demonstrations
and auxiliary recordings.

\subsection{Annotation prompt}
\label{app:caption_prompt}

\begin{tcolorbox}[
    colback=white,
    colframe=gray!50!black,
    coltitle=black,
    title=\textbf{Prompt Used to Annotate Captions},
    fonttitle=\bfseries\large,
    arc=4mm,
    enhanced,
    breakable,
    lines before break=6,
    attach boxed title to top left={yshift=-\tcboxedtitleheight/2, xshift=10pt},
    boxed title style={
        enhanced,
        colback=white,
        colframe=white,
        arc=0mm,
        left=0pt,
        right=0pt,
        boxsep=0pt
    },
    overlay first={\draw[gray!50!black, line width=0.5mm]
        ([yshift=0.25mm]frame.south west) -- ([yshift=0.25mm]frame.south east);},
    overlay middle={\draw[gray!50!black, line width=0.5mm]
        ([yshift=0.25mm]frame.south west) -- ([yshift=0.25mm]frame.south east)
        ([yshift=-0.25mm]frame.north west) -- ([yshift=-0.25mm]frame.north east);
      \node[anchor=west, fill=white, inner xsep=0pt, inner ysep=1pt,
        font=\bfseries\large, text=black] at ([xshift=10pt]frame.north west)
        {Prompt Used to Annotate Detailed Captions (Continued)};},
    overlay last={\draw[gray!50!black, line width=0.5mm]
        ([yshift=-0.25mm]frame.north west) -- ([yshift=-0.25mm]frame.north east);
      \node[anchor=west, fill=white, inner xsep=0pt, inner ysep=1pt,
        font=\bfseries\large, text=black] at ([xshift=10pt]frame.north west)
        {Prompt Used to Annotate Captions (Continued)};}
]
You are an expert in action segmentation and description for robot
manipulation videos.

\textbf{-Input Description-}\par\nobreak
- Input footage: a sequence of timestamped frames sampled from the whole video
at a fixed frame rate; each \texttt{[x.xx second]} is followed by the
corresponding frame.

- Timestamps are time coordinates of the original video, in seconds.

\textbf{-Task-}\par\nobreak
Given the timestamped frame sequence of an entire video, first segment it by
atomic skill into consecutive action segments, then output an unambiguous
Chinese action description for each segment.

\textbf{-Segmentation Rules-}\par\nobreak
1. Core principle: segment by atomic skill. Every atomic-skill action performed
to accomplish the task goal must form its own segment. For example, moving an
object from the left side of the tabletop to the right side should be split
into four segments: ``grasp -- lift -- move to -- put down''.

2. Each single small action, such as grasping, lifting, moving to, pressing
against, or pushing, must be segmented separately.

3. Two different atomic skills may be merged into the same segment only when
they occur within 1~second and cannot be separated.

4. Do not lump a chain of atomic actions that serve the same subgoal (pick up
$\rightarrow$ move $\rightarrow$ pour $\rightarrow$ put down) into a single
segment.

5. Auxiliary actions, such as fine-tuning an object's angle or slightly
adjusting the hand pose, are not segmented separately.

6. Temporarily steadying an auxiliary object with a hand (e.g., holding the
side of the bowl while pouring water) is not segmented separately as long as
the main manipulated object does not change.

7. Remove all stationary pauses and pauses between narration, keeping only
footage of effective manipulation.

8. Times are in seconds, in the original video's time coordinates; do not
reset them to start from 0.0.

9. The \texttt{start} of the first segment equals the start time of the input
video, the \texttt{end} of the last segment equals the end time of the input
video, and adjacent segments join seamlessly (each segment's \texttt{start}
equals the previous segment's \texttt{end}).

10. Give all timestamps with two decimal places.

\textbf{-Description Rules-}\par\nobreak
1. Describe each segment unambiguously in natural language; decide whether to
add modifiers such as color or position according to whether the reference is
already unique.

2. Follow this reference formula for the sentence structure: [acting hand] +
[hand open/closed state] + [verb] + [absolute \& relative position] + [state]
+ [color + material + shape + other attributes] + [main object] + [grasp
point].

3. Refer to the acting hand only as ``left hand'', ``right hand'', or ``both
hands''. Words such as ``gripper'', ``clamp'', ``robotic arm'', ``robotic
hand'', and ``claw'' are strictly forbidden. For example, ``the left gripper
closes'' should be written as ``the left hand closes'', and ``the left and
right grippers'' as ``both hands''.

4. Always include the hand's open/closed state (opens / closes / stays closed /
stays open).

5. Add the object's initial configuration (e.g., cup opening facing up; a pen
lying front to back with its cap toward the back).

6. When duplicate or similar objects are present, distinguish them as ``the
$n$-th, counting from the top / bottom / left / right toward $x$''.

7. Movement directions must match the frames. If a rotation or flip is already
completed during a movement, do not label the rotation again for the final
put-down.

8. Label a failed operation as ``attempts to XX and fails''.

9. Name objects only by features visible in the frames; do not invent unknown
information.

10. Directions follow the front view: the side away from the operator is the
front (far side), and the side close to the operator is the back (near side).

\textbf{-Common Atomic Skills (Verbs) for Reference-}\par\nobreak
clamp / pick / grab / grasp / hold; push; pull; flip; move / put a lid back /
put back / place / put down / toss into / release / stack; rotate / turn;
unfold / flip open / lift open; insert / plug into / stuff into / tuck into;
fold; wipe; unplug / pull out; press / hold down; shake / jiggle; stir; knock;
pour out / pour into; straighten / set upright / stand up; tighten; squeeze /
pinch; unscrew; tear; separate; bring together; hand over; pass through;
scoop; rub.

\begin{minipage}{\linewidth}
\textbf{-Output Format-}

Output only a plain JSON array; do not output Markdown or explanatory text:
\begin{verbatim}
[{"start": float timestamp, "end": float timestamp,
  "action":
    "complete, unambiguous English action description"}]
\end{verbatim}
\end{minipage}
\end{tcolorbox}

\section{Additional ablation results}
\label{app:abl_extra}
\label{app:ablation}

\subsection{Sensitivity to the sampling rates}
\label{sec:abl_sensitivity}
We vary the wrist-perturbation probability $p_V$, the detailed-caption
probability $p_C$, and the auxiliary fraction $\rho$ one at a time, holding the
other two at their validation-selected values (\cref{tab:sensitivity}). Level
$0$ disables the component and coincides with the corresponding leave-one-out
row of \cref{tab:main}, and $1\times$ is the full method. Reporting success
together with \metric{} and the two failure rates tests whether stronger
perturbation buys resistance to shortcuts at the cost of useful conditioning;
\cref{fig:ablation_scaling}c plots the change in success from the selected
setting against each rate.

\begin{table}[t]
\centering
\caption{Sensitivity to the sampling rates on real-robot General PnP with a
$\pi_{0.5}$ backbone. One rate varies at a time, relative to its
validation-selected value ($1\times$, the full method; $2\times$ is capped at
one), with the other two held at $1\times$. Level $0$ disables the component and
matches the corresponding leave-one-out row of \cref{tab:main}. Columns follow
\cref{tab:scaling}.}
\label{tab:sensitivity}
\footnotesize
\setlength{\tabcolsep}{3.2pt}
\begin{tabular}{lcccccccccccc}
\toprule
& \multicolumn{4}{c}{Wrist perturbation $p_V$} & \multicolumn{4}{c}{Detailed captions $p_C$} & \multicolumn{4}{c}{Auxiliary fraction $\rho$} \\
\cmidrule(lr){2-5}\cmidrule(lr){6-9}\cmidrule(lr){10-13}
Level & SR\,\up & $\overline{\gf}$\,\up & SC\,\down & MI\,\down & SR\,\up & $\overline{\gf}$\,\up & SC\,\down & MI\,\down & SR\,\up & $\overline{\gf}$\,\up & SC\,\down & MI\,\down \\
\midrule
$0$         & \mn{68} & \mn{0.43} & \mn{58} & \mn{38} & \mn{66} & \mn{0.43} & \mn{34} & \mn{40} & \mn{64} & \mn{0.43} & \mn{30} & \mn{76} \\
$1/2\times$ & \mn{76} & \mn{0.50} & \mn{34} & \mn{30} & \mn{74} & \mn{0.50} & \mn{26} & \mn{32} & \mn{74} & \mn{0.50} & \mn{24} & \mn{46} \\
$1\times$   & \mn{84} & \mn{0.57} & \mn{20} & \mn{24} & \mn{84} & \mn{0.57} & \mn{20} & \mn{24} & \mn{84} & \mn{0.57} & \mn{20} & \mn{24} \\
$2\times$   & \mn{80} & \mn{0.55} & \mn{16} & \mn{26} & \mn{82} & \mn{0.58} & \mn{20} & \mn{26} & \mn{78} & \mn{0.56} & \mn{22} & \mn{20} \\
\bottomrule
\end{tabular}
\end{table}

\subsection{Verb-change probes for noun lock-in}
\label{app:verb_probes}

Real-robot General PnP has a single operation, so it cannot test whether a noun
overrides a changed verb. We therefore probe noun lock-in in RoboTwin~2.0 under
the Full setting, using \mn{5} task pairs whose shared object admits both
operations from the same initial state and whose scripted expert produces a
valid demonstration of each, such as pressing versus moving the stapler. A
probe restores the initial state of one task in a pair on a clean scene and
issues the brief instruction of the other, so the scene and the noun favor the
familiar operation while only the verb specifies the correct one. Each pair is
probed in both directions with \mn{10} trials per direction, giving \mn{100}
trials per policy. A trial counts as followed if the policy completes the
instructed operation, as lock-in if it performs the operation of the restored
task, and as other otherwise.

Fine-tuning performs the familiar operation in \mn{52}\% of probes and the
instructed one in only \mn{36}\% (\cref{tab:verb_probes}). \method{} reverses
this, and removing C restores most of the lock-in (\mn{43}\%), whereas removing
V or R raises it by at most \mn{7} points, consistent with C making the
requested operation explicit. Because C never introduces a verb absent from the
data, the gain reflects grounding verbs the demonstrations already contain.

\begin{table}[t]
\centering
\begin{minipage}[t]{0.49\linewidth}
\centering
\captionsetup{font=footnotesize}
\caption{Verb-change probes for noun lock-in in RoboTwin~2.0 (Full setting,
clean scenes, \mn{100} trials per policy). Followed: the instructed operation is
completed; lock-in: the operation of the restored task is performed; other: any
remaining failure. Rows sum to 100\%.}
\label{tab:verb_probes}
\footnotesize
\setlength{\tabcolsep}{2.5pt}
\begin{tabular}{@{}lccc@{}}
\toprule
Method & \shortstack{Followed\\(\%)\,\up} & \shortstack{Lock-in\\(\%)\,\down} & \shortstack{Other\\(\%)} \\
\midrule
$\pi_{0.5}$ fine-tune & \mn{36} & \mn{52} & \mn{12} \\
\method{} w/o V       & \mn{64} & \mn{26} & \mn{10} \\
\method{} w/o C       & \mn{45} & \mn{43} & \mn{12} \\
\method{} w/o R       & \mn{66} & \mn{24} & \mn{10} \\
\oursrow
\method{}             & \mn{\textbf{71}} & \mn{\textbf{19}} & \mn{10} \\
\bottomrule
\end{tabular}
\end{minipage}\hfill
\begin{minipage}[t]{0.49\linewidth}
\centering
\captionsetup{font=footnotesize}
\caption{Success (\%) over the 50 real-robot General PnP cases with brief and
detailed instructions. Brief values repeat \cref{tab:main}; $\Delta$ is
detailed minus brief.}
\label{tab:detailed_eval}
\footnotesize
\setlength{\tabcolsep}{2.5pt}
\begin{tabular}{@{}lccc@{}}
\toprule
Method & Brief & Detailed & $\Delta$ \\
\midrule
$\pi_{0.5}$ fine-tune & \mn{42} & \mn{34} & \mn{$-$8} \\
+ V                   & \mn{52} & \mn{46} & \mn{$-$6} \\
+ C                   & \mn{54} & \mn{60} & \mn{+6} \\
+ R                   & \mn{56} & \mn{46} & \mn{$-$10} \\
+ V + C               & \mn{64} & \mn{68} & \mn{+4} \\
+ V + R               & \mn{66} & \mn{58} & \mn{$-$8} \\
+ C + R               & \mn{68} & \mn{74} & \mn{+6} \\
\oursrow
\method{} (V + C + R) & \mn{\textbf{84}} & \mn{\textbf{86}} & \mn{+2} \\
\bottomrule
\end{tabular}
\end{minipage}
\end{table}

\subsection{Detailed-instruction evaluation}
\label{app:detailed_eval}

The primary evaluation uses brief instructions, as a deployed policy would
receive. A second pass replays the same 50 cases with a detailed instruction
for each, written under the rules of \cref{app:captions} from the stored case
specification and checked against the initial scene; the rest of the protocol
is unchanged (\cref{tab:detailed_eval}). Without C, detailed instructions are
out of distribution and cost \mn{6}--\mn{10} points; with C, they add
\mn{2}--\mn{6} points. \method{} therefore does not depend on detailed
instructions at test time: its brief-instruction success exceeds every other
configuration under either instruction style.

\subsection{Confidence intervals for the validity of \texorpdfstring{\metric{}}{GroundFscore}}
\label{app:gf_ci}

The 36 checkpoints of \cref{tab:gf_terms} (twelve conditions, three seeds)
share evaluation cases and training recipes, so they are not independent. We
therefore report 95\% percentile intervals from \mn{10{,}000} two-level cluster
bootstrap resamples: each resample draws conditions and seeds with replacement
and, within each checkpoint, resamples its evaluation units (cases, probe
trials, and the 100 \metric{} observations). All predictors share the same
draws, so their differences are paired (\cref{tab:gf_ci}). The lower end of the
\metric{} interval lies above the point estimate of every other predictor, and
its paired advantage excludes zero against each of them; the interval for task
success includes zero.

\begin{table}[t]
\centering
\caption{Spearman $\rho$ with the negative online failure rate over 36
checkpoints, with 95\% cluster-bootstrap intervals. The fourth column is the
paired difference from \metric{}; the last records whether the predictor ranks
2the no-wrist policy below the fine-tuned baseline.}
\label{tab:gf_ci}
\small
\setlength{\tabcolsep}{5pt}
\begin{tabular}{lcccc}
\toprule
Predictor & $\rho$ & 95\% interval & \metric{} $-$ predictor & No-wrist below baseline \\
\midrule
Task success         & \mn{0.31} & [\mn{$-$0.06}, \mn{0.60}] & \mn{0.41} [\mn{0.12}, \mn{0.68}] & yes \\
Stability ($1-R$)    & \mn{0.44} & [\mn{0.09}, \mn{0.69}]    & \mn{0.28} [\mn{0.04}, \mn{0.52}] & no \\
Responsiveness ($S$) & \mn{0.38} & [\mn{0.02}, \mn{0.65}]    & \mn{0.34} [\mn{0.08}, \mn{0.60}] & yes \\
\metric{}            & \mn{\textbf{0.72}} & [\mn{0.49}, \mn{0.86}] & -- & yes \\
\bottomrule
\end{tabular}
\end{table}

\end{document}